# A MICROWAVE IMAGING AND ENHANCEMENT TECHNIQUE FROM NOISY SYNTHETIC DATA

[1]*ANJAN KUMAR KUNDU, [2] BIJOY BANDYOPADHYAY, [3]SUGATA SANYAL,

[1] INSTITUTE OF RADIOPHYSICS & ELECTRONICS, UNIVERSITY OF CALCUTTA,
92 A.P.C ROAD, KOLKATA, INDIA.
E-mail: akkundu.rpe@caluniv.ac.in

[2] INSTITUTE OF RADIOPHYSICS & ELECTRONICS, UNIVERSITY OF CALCUTTA,
92 A.P.C ROAD, KOLKATA, INDIA.
E-mail: bb.rpe@caluniv.ac.in

[3] TATA INSTITUTE OF FUNDAMENTAL RESEARCH, INDIA
E-mail: sanyal@tifr.res.in

**ABSTRACT:**

An inverse iterative algorithm for microwave imaging based on moment method solution is presented here. The iterative scheme has been developed on constrained optimization technique and is certain to converge. Different mesh size for the model has been used here to overcome the Inverse Crime. The synthetic data at the receivers is contaminated with different percentage of noise. The ill-posedness of the problem is solved by Levenberg - Marquardt method. The algorithm is applied to synthetic data and the reconstructed image is then further enhanced through the Image enhancement technique

## 1. INTRODUCTION

For the last few decades, microwave tomography techniques for biomedical applications have been received increasing interest. Intensive studies in this field able to give an efficient solution to quantitative imaging. We had proposed several algorithms [1-2] which though reconstructed the image without any misfit, yet the mesh size remains the same both in the forward and the inverse problem leading to inverse crime. This paper represents an iterative algorithm based on Levenberg-Marquardt regularization technique with necessary considerations to avoid inverse crime. The reconstructed image is then undergone through image enhancement mechanisms to minimize the noise.

## 2. THE STUDY

During the past 20 years, immense research has been carried out in microwave tomography to quantitatively reconstruct the complex permittivity distribution of the biological media. Standard spectral diffraction tomography [3-7] which has been investigated with application to microwaves, prove to be very fast and capable of producing reconstructions with good quantitative accuracy for small contrast objects.

∗ Corresponding Author

Yet it is subjected to various limitations, including the artifacts due to the diffraction effects in strongly inhomogeneous media where Born or Rytov approximations are not valid [8-9].

Several approaches based on moment methods [10-12] have, in past, been explored rigorously, but the stability depends on the measurement accuracy due to ill-conditioning of the matrix. Also the convergence depends on the contrast of the objective.

In recent years, multiplicative regularized contrast source inversion method is applied to microwave biomedical applications [13-14]. The inversion method is fully iterative and avoids solving any forward problem in each iterative step.

In our earlier works [15-16], we had suggested quasi-ray optic SIRT-style algorithms for microwave imaging. In those first generation algorithms it was assumed that only those cells situated within the beamwidth of the transmitter radiation pattern, would effectively contribute to the field at the end of a ray. Those linear and nonlinear algorithms did not reconstruct the image quantitatively to the extent which could be considered to be clinically important.

The earlier algorithms proposed by us [1-2] which, though reconstructed the image without any misfit, yet the same mesh size for both in the forward and the inverse problem leading to inverse crime. This paper represents an iterative algorithm with necessary considerations to avoid inverse crime.

## 3. ANALYSIS, DISCUSSIONS, APPROACHES AND INTERPRETATIONS

### 3.1. FORWARD PROBLEM

The forward problem has already been discussed in details in our previous work [2]. A cylindrical object of arbitrary cross section is considered here which is characterized by a complex permittivity distribution $\varepsilon(x,y)$. An electromagnetic wave radiated from an open-ended waveguide is used here for the illumination. The incident electric field $E^{inc}$ is parallel to the axis of the cylinder.

The expression for the total electric field E is

$$\vec{E} = \vec{E}^{inc} + \vec{E}^{s} \qquad (1)$$

where $E_s$ represents the scattered field which is generated by the equivalent electric current radiating in free space.

The total electric field can be calculated with an integral representation

$$\vec{E}(x, y) = \vec{E}^{inc}(x, y) + \int\int_s J(x, y) G(x, y; x', y') \, dx' dy' \qquad (2)$$

where the Green's function can be given by

$$G(x, y; x', y') = -\frac{j}{4} H_0^2 \left( k\sqrt{(x-x')^2 + (y-y')^2} \right) \qquad (3)$$

Here $(x, y)$ and $(x', y')$ are the observation and source points respectively.

The solution of the forward problem are carried out by moment method [17] using pulse-basis function and point matching technique. The synthetic data at the receivers is then contaminated with different percentage of noise as our main objective is to reconstruct the numerical model under noisy conditions.

## 3.2. INVERSE PROBLEM

The aim of the inverse problem is to find a stable solution for permittivity distribution $\varepsilon^*$ which minimizes the squared error output at the receivers i.e.

$$\|E(\varepsilon) - e\|_2^2 \quad (4)$$

where

$e \in C^n$, the n electric fields we measure at receiver points,

$E : C^m \to C^n$, a function mapping the complex permittivity distribution with m degrees of freedom into a set of n approximate electric field observations,
and also

$\varepsilon \in C^m$, the complex permittivity distributions with m degrees of freedom.

The Levenberg-Marquardt regularization technique for the minimization of the (4) leads to an iterative solution

$$\varepsilon_{i+1} = \varepsilon_i + \Delta \varepsilon_i \quad (5)$$

where
$\varepsilon_{k+1}$ is the permittivity distribution at the k+1$^{th}$ iteration.
$\Delta \varepsilon$ can be written as

$$\Delta \varepsilon = (E'(\varepsilon)^\dagger E'(\varepsilon) + \lambda I)^{-1} E'(\varepsilon)^\dagger (E(\varepsilon) - e) \quad (6)$$

where $E'$ is the Jacobian matrix, $\dagger$ denotes the conjugate transpose, $\lambda$ is a monotonically decreasing regularization parameter, I is the identity matrix, $E(\varepsilon)$ is the calculated electric fields at the receivers.

## 3.3. NUMERICAL MODEL

To test our algorithm, we have considered the following model as shown in Figure 1.

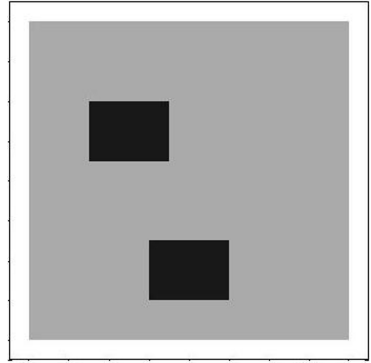

**Figure 1.** Numerical model

It is a high contrast square biological object 9.6 cm × 9.6 cm consisting of muscle and bone having complex dielectric constants 50-j23 and 8-j1.2 respectively at a frequency of 1 GHz. The object is kept immersed in saline water having complex dielectric constant 76-j40.

The target is illuminated with TE fields radiating from an open ended dielectric filled wave guide having sinusoidal aperture field distribution. The transmitter is moved along four mutually orthogonal directions. For each of the transmitter positions along a particular transmitting plane, the received fields at eighteen locations in the other three orthogonal planes were measured theoretically at a frequency of 1 GHz

Different meshes are used to overcome the inverse problem as shown in figure 2. The finer mesh is used in the forward problem (Figure 2(a)) whereas the inverse solver uses the coarse mesh (Figure 2 (b)).

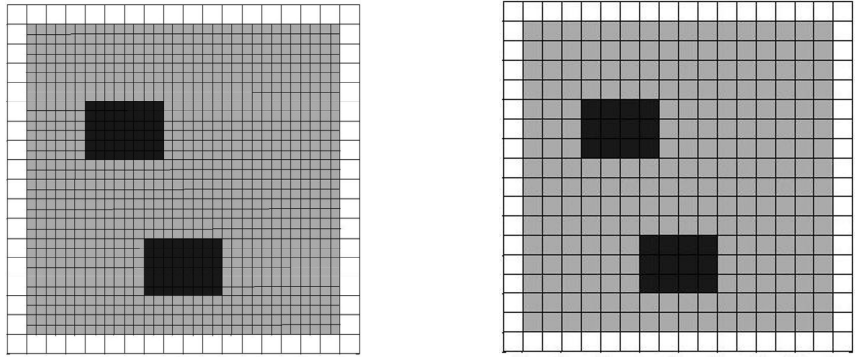

**Figure 2.** Meshes used to overcome the inverse crime (a) Mesh used in forward problem (b). Mesh used in inverse problem

In case of forward problem, the rectangular model is divided into 1024 square cells of dimension 0.3cm X 0.3cm and the saline water region is divided into 32 cells of dimension 0.6cm X 0.6cm. During the inverse problem, the rectangular model together with saline water region is divided into 324 equal square cells 0.6 cm× 0.6 cm. The measurement set contains 288 independent data.

During the iterative reconstruction, the complex permittivity values of the cells filled up with saline water were assumed to be known, thus rendering the problem of estimating the complex dielectric constants of the remaining 256 cells.

### 3.4. RESULTS AND DISCUSSIONS

To apply the reconstruction algorithm, it was initially assumed that the biological medium is filled up with muscle only. The received fields at different receiver locations were computed for each transmitter position. The only priori information we have used in our algorithm is that the real part of the complex dielectric constant cannot be negative and the imaginary part cannot be positive. The iteration is stopped when the 2-norm error output is of the order of $10^{-4}$.

The reconstructed models for different percent of noise in the synthetic data are then undergone through image enhancement filter in particular through the histogram equalization technique to have an improved quality of reconstructed model in terms of noise. The figures 3 to 10 display the reconstructed models for different percent of noise and their enhancement in quality after using the Image adjustment technique.

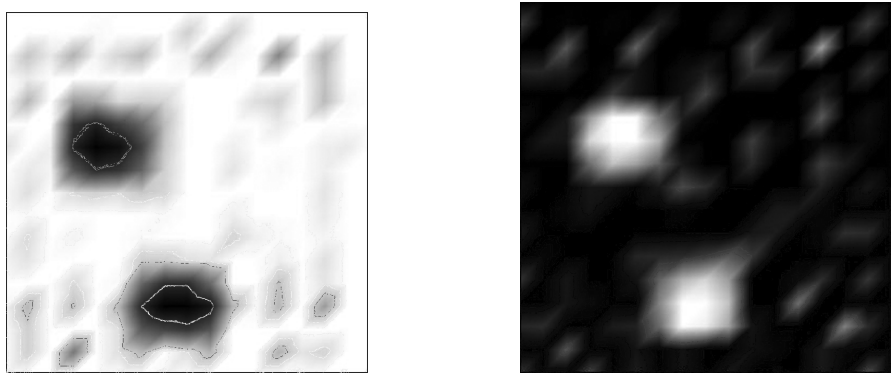

**Figure 3.** Reconstructed model with 1%noise (a) Real part (b) Imaginary part

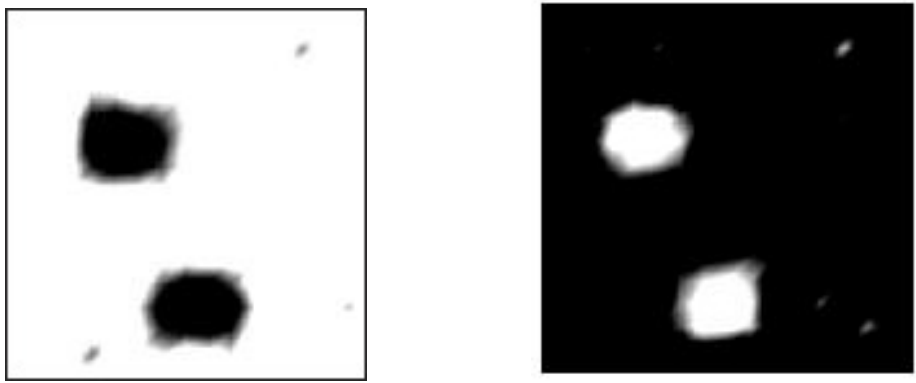

**Figure 4.** Reconstructed model with 1%noise after using image adjustment technique (a) Real part (b) Imaginary part

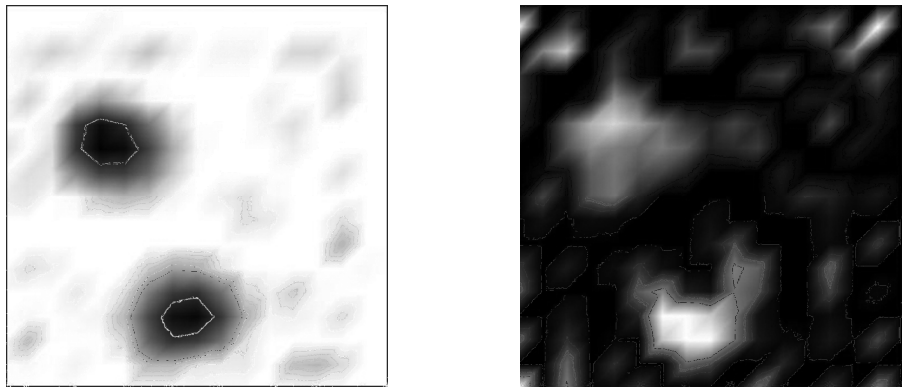

**Figure 5.** Reconstructed model with 2%noise (a) Real part (b) Imaginary part

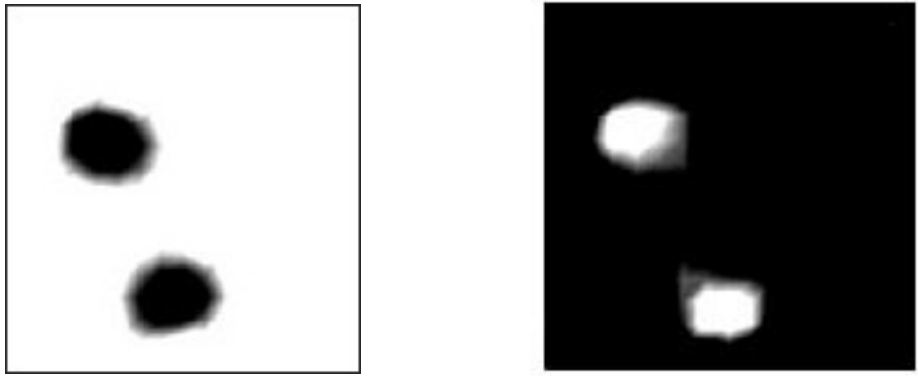

**Figure 6.** Reconstructed model with 2%noise after using image adjustment technique (a) Real part (b) Imaginary part

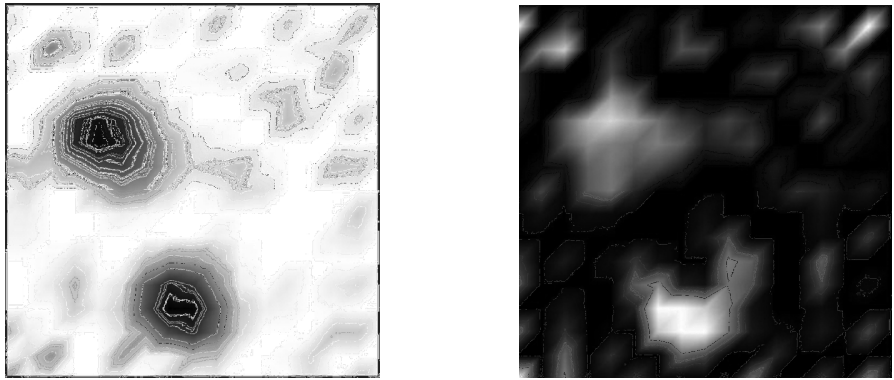

**Figure 7.** Reconstructed model with 5%noise (a) Real part (b) Imaginary part

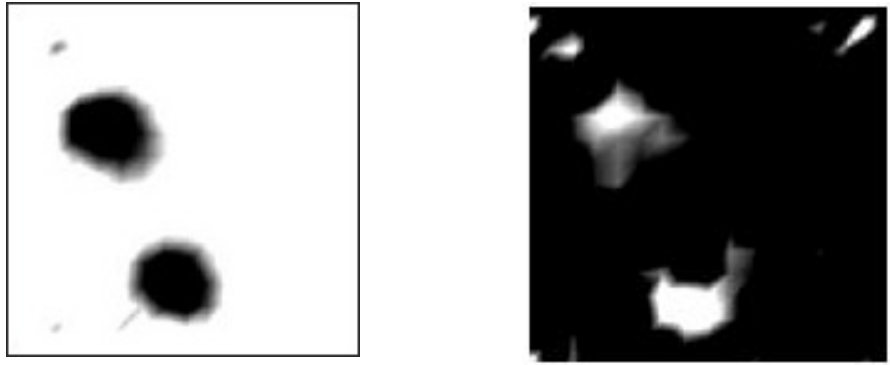

**Figure 8.** Reconstructed model with 5%noise after using image adjustment technique (a) Real part (b) Imaginary part

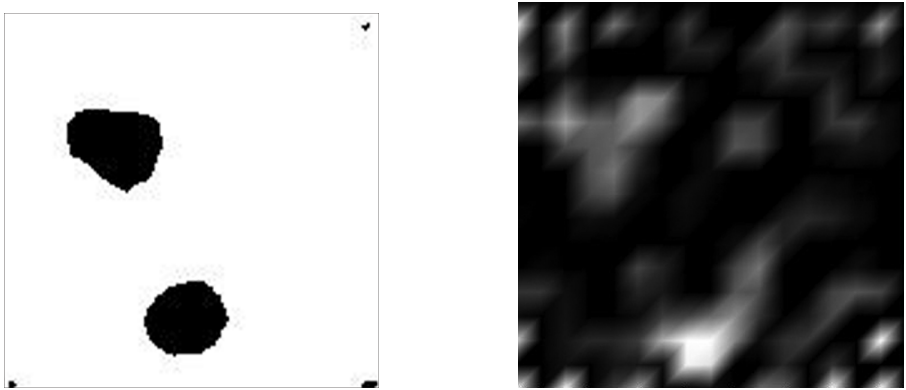

**Figure 9.** Reconstructed model with 10%noise (a) Real part (b) Imaginary part

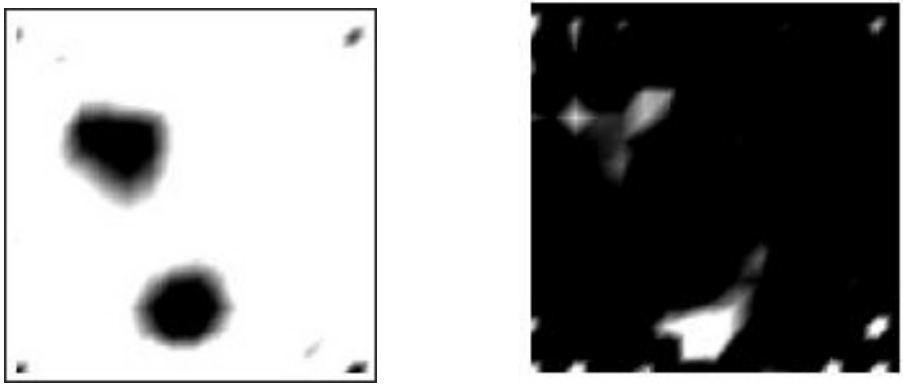

**Figure 10.** Reconstructed model with 10%noise after using image adjustment technique (a) Real part (b) Imaginary part

## 4. CONCLUSIONS

Thus using the different mesh sizes in the forward and reverse problem, the inverse crime has been avoided in this proposed algorithm. Also the clarity of the reconstructed images for different percentage of noise has been improved through Image adjustment technique. This work can be further extended by incorporating different other regularization techniques along with other image enhancement techniques for a comparative study